# BaRT: A Bayesian Reasoning Tool for Knowledge Based Systems

*Lashon B. Booker, Naveen Hota\*, and Connie Loggia Ramsey*

Navy Center for Applied Research in Artificial Intelligence
Code 5510
Naval Research Laboratory
Washington, DC 20375-5000

## Abstract

As the technology for building knowledge based systems has matured, important lessons have been learned about the relationship between the architecture of a system and the nature of the problems it is intended to solve. We are implementing a knowledge engineering tool called BaRT that is designed with these lessons in mind. BaRT is a Bayesian reasoning tool that makes belief networks and other probabilistic techniques available to knowledge engineers building classificatory problem solvers. BaRT has already been used to develop a decision aid for classifying ship images, and it is currently being used to manage uncertainty in systems concerned with analyzing intelligence reports. This paper discusses how state-of-the-art probabilistic methods fit naturally into a knowledge based approach to classificatory problem solving, and describes the current capabilities of BaRT.

## 1. Introduction

As the technology for building knowledge based systems has matured, many important lessons have been learned about how to apply this technology to real world problems. One of the most important of these lessons is that methods for representing and manipulating uncertainty must be considered an integral part of the overall design. Early system designers mistakenly assumed that uncertainty could simply be "added on" to rule based representations, treating uncertain inferences just like logical ones. The essential idea was that uncertain inferences could always be modularized. An inference of the form $A \Longrightarrow B$ is modular in the sense that belief in $B$ is updated no matter how our belief in $A$ was derived, and no matter what else is in the knowledge base (Pearl, 1988). This point of view has recently been questioned, however, as the implications of the modularity assumption are more clearly understood (Henrion, 1986;Heckerman & Horvitz, 1987;Pearl, 1988). The problem is that uncertain reasoning often must handle dependencies among hypotheses that are inherently not modular. Accounting for these dependencies requires reasoning capabilities which can be difficult to implement using a collection of modular rules. An alternative approach to this problem is to abandon the modularity of rule-based updating and represent the relationships among hypotheses explicitly. This type of representation is available in computational schemes that use *belief networks* of various kinds (Pearl-86; Shachter, 1988; Shenoy & Shafer, 1986). Belief networks provide an economical summary of the relationships among hypotheses as well as an axiomatic computation of uncertainty.

In addition to their systematic treatment of uncertainty, belief networks also comply with other lessons learned about knowledge based systems. For example, it is now widely agreed that inference strategies should be tailored to work well with the knowledge representations they reason about; and, that a causal or functional model of a problem can provide more reasoning power than a collection of empirical associations (Davis, 1982). Belief networks provide a qualitative model of the inherent causal structure of a problem in uncertain reasoning. The many dependencies and implicit relations that must be listed exhaustively in a rule-based approach are efficiently summarized by the paths between nodes in a belief network. Moreover, belief networks can be used as inference engines. The information needed to update the belief distribution at a node is available locally from that node's neighbors. This makes it possible to use distributed, message-passing computations to propagate the effects of changes in belief. Since the computation only examines interactions among semantically related variables, each step in the process has a meaningful interpretation.

---

\* Author is employed by JAYCOR, 1608 Spring Hill Road, Vienna, VA 22180

Because of these many advantages, it is not surprising that software tools have been developed to make belief networks and related techniques more accessible to knowledge engineers (eg. (Shachter, 1988), (Chavez & Cooper, 1988)). While these tools have already proven to be useful in many applications, this paper contends that they do not go far enough in meeting the requirements of current knowledge based systems. In particular, these tools force the user to represent all of the knowledge relevant to a problem in the same knowledge representation. The use of a single uniform knowledge representation is incompatible with two current trends in knowledge based system design: the use of multiple specialized representations for each kind of knowledge (Davis, 1982), and an overall system architecture that integrates these representations in a way that reflects the inherent structure of the problem being solved (Chandrasekaran, 1986). The Navy Center for Applied Research in AI (NCARAI) is implementing a knowledge engineering tool called BaRT that is designed with these lessons in mind.

BaRT is a Bayesian reasoning tool that makes belief networks and other probabilistic techniques available to knowledge engineers building classificatory problem solvers. This paper discusses how state-of-the-art probabilistic methods fit naturally into a knowledge based approach to classificatory problem solving, and describes the current capabilities of BaRT.

## 2. Classificatory Problem Solving

Because many knowledge based systems accomplish some form of classification, a considerable amount of work has been done to analyze classificatory problem solving as a generic phenomenon (Clancey, 1984; Chandrasekaran, 1986). Such an understanding can provide insights about the kinds of knowledge structures and control regimes that are characteristic of these problem solving tasks in general. This, in turn, allows researchers to develop system building tools that reflect the inherent structure of classificatory problem solving, facilitating system design, knowledge acquisition, and explanation. In this section we briefly discuss the generic view of classificatory problem solving and describe how uncertainty calculations play a crucial role.

### 2.1. Generic tasks

Chandrasekaran (1986) argues that the inherent structure of any problem solving task can be revealed by decomposing it into elementary organizational and information processing strategies called *generic tasks*. A generic task analysis of classificatory problem solving can be gleaned from the basic elements shown in Figure 1. Chandrasekaran uses the term *hierarchical classification* to describe the primary generic task associated with classificatory problem solving. Hierarchical classification inherently involves selecting from a pre-enumerated set of solutions, hierarchically organized in terms of class-subclass relationships. The problem solving process resolves the impact of evidence, and establishes which hypotheses are most strongly confirmed. Those hypotheses are then refined by gathering additional evidence, when available, that might confirm or refute hypotheses lower in the hierarchy. In this way, the process tries to establish the most specific solution possible.

In very simple situations, hierarchical classification is the only generic task required for classificatory problem solving. In most cases, though, the extent to which the available data constitutes evidence for establishing or rejecting a hypothesis cannot be known directly and must be inferred. Two kinds of generic tasks are available to support such inferences. The first is called *knowledge-directed information passing*. This task accomplishes simple categorical inferences that exploit conceptual relationships in the data. Examples of these inferences include definitional abstractions (eg. x lower than y implies y is higher than x), database retrieval, and simple property inheritance.

When the data available is uncertain, incomplete or only partially matches a hypothesis, straightforward categorical inferences are no longer adequate. The fit between data and hypotheses must then be determined by another generic task called *hypothesis matching*. As the name implies, hypothesis matching is the process of computing the extent to which evidence confirms, rejects or suggests a hypothesis. The emphasis is on managing the uncertainty associated with both the data and the knowledge needed to relate it to the hypotheses of interest. Several familiar methods are available for implementing this task: statistical pattern classification algorithms, inference networks, heuristic rules, and so on.

This generic view of classificatory problem solving makes it clear that several kinds of inferences and relationships among hypotheses might be required. In the hierarchical classification task, evidence must be combined at a level of abstraction commensurate with the available data, in a manner consistent with the object-class relationships in the hierarchy. The knowledge-directed information passing task could require a set of logical rules to characterize property inheritance relationships. Hypothesis matching might involve inferences based on object-part relationships, causal relationships, geometric relationships, or temporal relationships. Since the reasoning associated with each task may be hierarchical, the overall classification process can involve several levels of abstraction, each perhaps having its own reasoning mechanism, all integrated by the primary hierarchical classification task.

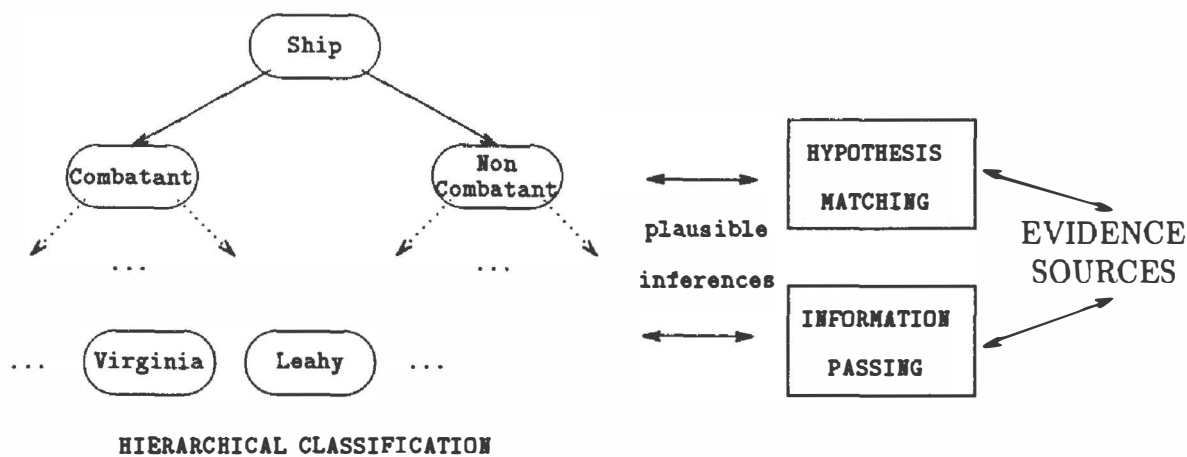

Figure 1: A generic view of classificatory problem solving

Note that the need to refine confirmed hypotheses points to an important control problem: deciding when additional evidence should be acquired and processed, and deciding which inferences should be computed. The efficiency of classificatory problem solving obviously depends on making cost-effective use of both the available evidence and computational resources.

### 2.2. The CSRL Approach

The Conceptual Structures Representation Language (CSRL) (Bylander & Mittal, 1986) is designed to facilitate the construction of classificatory problem solvers in a way commensurate with the generic task point of view. Language constructs are provided for all aspects of classificatory problem solving: defining classification hierarchies, delineating problem solving strategies, and calculating a degree of certainty for each classificatory hypothesis.

The basic unit of representation in CSRL is the *specialist* that designates a particular hypothesis or concept related to the problem. Each specialist is a knowledge based problem solving agent that can make local decisions about how well it fits the data available. Once a specialist has been established as a relevant hypothesis, it can decide which other specialists should be invoked. Specialists in CSRL are organized into a classification tree reflecting the class-subclass relationships most important to the problem. Hierarchical classification proceeds according to a hypothesis refinement scheme called *establish-refine*. When a hypothesis is established as relevant, it refines itself by invoking its subhypotheses and other hypotheses indicated by the data so that they can try to establish themselves. Specialists communicate with each other by passing messages, and each specialist has local procedures specifying how it responds to each kind of message.

Uncertainty calculations in CSRL are used to determine when a hypothesis has been established. The key unit of representation here is the *knowledge group*. A knowledge group is a collection of production rules that map problem data into a discrete scale of qualitative confidence values. This mapping implements either the hypothesis matching task, or the knowledge-directed information passing task. Each knowledge group corresponds to an evidential abstraction needed to establish a hypothesis. A hypothesis is established once it achieves certain distinguished levels of confidence. The evidence used by a knowledge group to make this determination can range from low level raw data or database

48

queries to the abstract evaluations made by other knowledge groups. This gives a hierarchical organization to the uncertainty calculation. Hierarchical relationships among knowledge groups correspond to levels of conceptual abstraction, and the informational dependencies between knowledge groups correspond to the evidential relationships underlying a classificatory hypothesis.

Two points about the CSRL framework should be noted. First, each classificatory hypothesis is associated with its own hierarchy of knowledge groups dedicated to computing its confidence value. Each knowledge group is therefore a specialized body of knowledge that is brought to bear only when its associated hypothesis tries to establish itself. Moreover, whenever a hypothesis is rejected, all of its subhypotheses are also ruled out along with their knowledge group hierarchies. This is an important way to exercise domain dependent control over the problem solving process. Second, it is important to note that qualitative scales are used to measure confidence levels. The basic premise is that the conditions for applying most numeric uncertainty calculi usually do not hold in practice (Chandrasekaran & Tanner, 1986). Translating expert knowledge into a *normative* calculus therefore becomes a step that unnecessarily introduces uncertainty into the problem solver.

The CSRL approach attempts to model as closely as possible the judgements and conceptual structures used by a human expert. CSRL hopes to avoid the need for normative methods by using the inherent organization of the problem solving task to make the determination of plausible solutions tractable. There is no experimental evidence, however, that the qualitative uncertainty methods of CSRL are really a better descriptive model of human judgement than numeric methods. Moreover, no theory is provided to justify the way confidence values are combined or to give confidence values a clear semantic interpretation.

### 2.3. Managing Uncertainty

The CSRL qualitative approach to uncertainty management assumes that plausible inferences do not have to be precisely correct or complete (Bylander & Chandrasekaran, 1987). Indeed, correctness and completeness are viewed as issues that can detract from more important considerations like choosing a reasoning strategy that matches the problem at hand, or determining which inferences are really important in a given situation. Bylander and Chandrasekaran (1987, p. 242) argue that

> In diagnosis, for example, there is much more to be gained by using abduction ...than by independently calculating the degree of certainty of each hypothesis to several decimal places of accuracy.

While we agree that the generic task idea is a powerful construct for organizing problem solving and domain knowledge, we reject the implicit assertion that this approach is incompatible with normative methods for calculating uncertainty.

State of the art approaches to probabilistic inference offer a wide range of representations and problem solving capabilities (Pearl, 1988; Shachter, 1988; Spiegelhalter, 1986). Pearl (1987), for example, has shown how a Bayesian belief network can efficiently compute a categorical interpretation of a body of evidence that constitutes a composite explanatory hypothesis accounting for all observed evidence. This technique is one way to implement the abductive task that provides a clear semantics for what is computed. Similarly, there is no reason why a probabilistic inference has to be complete and exhaustive. In fact, decision theory can be used as a framework for rigorously and explicitly identifying the factors underlying a decision about what to compute (Horvitz, Breese, & Henrion, 1988).

The CSRL approach also seems to underestimate the importance of special constraints on classificatory inferences that might arise in a given domain. In the kinds of target classification problems the Navy is interested in, for instance, the problem solving environment is unstructured in the sense that there is often little control over what evidence will be available or when it will become available (Booker, 1988). Control strategies must therefore be flexible and opportunistic. Work on a target hypothesis must be suspended if the evidence needed to definitively establish or reject it is not available, or if evidence arrives that makes some other hypothesis more attractive. Uncertainty management schemes that insist on relatively rigid strategies like establish-refine are not well suited to handle this situation. Target classification problems also tend to involve many inherently ambiguous relationships between features and objects. This gives rise to some subtle dependencies among hypotheses about those objects. Maintaining consistent beliefs for several explanations of a given feature requires careful attention to the way alternative causal hypotheses interact. Most rule-based formalisms for uncertain inference either handle this problem awkwardly or cannot handle it at all (Henrion, 1986).

Given the many kinds of inferences involved in classificatory problem solving, it is especially im-portant that uncertain beliefs about hypotheses are managed with representations and techniques based on explicit assumptions, sound theory, and clear semantics. Otherwise there is little reason to expect that the disparate belief computations can be combined into a coherent result. Moreover, if the various reasoning activities are to be orchestrated into an effective problem solving process, there must be some framework available that allows beliefs to be systematically converted into decisions that take cost-benefit considerations into account.

## 3. A Bayesian Reasoning Tool

The major thrust of target classification research at NCARAI is the design and implementation of a tool for hierarchical Bayesian reasoning called BaRT (Booker, 1988; Hota, Ramsey, & Booker, 1988). BaRT facilitates the construction of knowledge based systems that reflect the generic structure of classificatory problem solving. Just as important, BaRT computes axiomatic and normative inferences. In this section we describe BaRT's capabilities and indicate how these capabilities fulfill the requirements of classificatory problem solving.

### 3.1. Overall System Architecture

Figure 2 shows the overall BaRT architecture. There are three major components of BaRT: the knowledge acquisition system, the network compiler, and the core inference routines. The knowledge acquisition system has capabilities similar to those found in other interactive tools of this kind (eg. KNET (Chavez & Cooper, 1988), and DAVID (Shachter, 1986)). It takes advantage of the network structure to focus knowledge acquisition on a single node at a time, quantifying the relationship between that consequent node and its immediate antecedents. BaRT also provides a collection of canonical descriptions of probabilistic interactions that the user can instantiate for any node in the network. It is important that the user is not forced to use the same canonical model to describe all joint interactions. BaRT currently offers the standard "noisy OR" (Henrion, 1986) and "noisy AND", generalized to be used with non-binary variables by identifying dominance relationships associated with the consequent node (Kim, 1983). Work is underway to include conjunctive and disjunctive models for binary variables that specify patterns of dependent relationships (Bonissone & Decker, 1986). Note that BaRT allows users to maintain a library of their own predefined subnetworks and canonical interactions that can be described once and stored until needed.

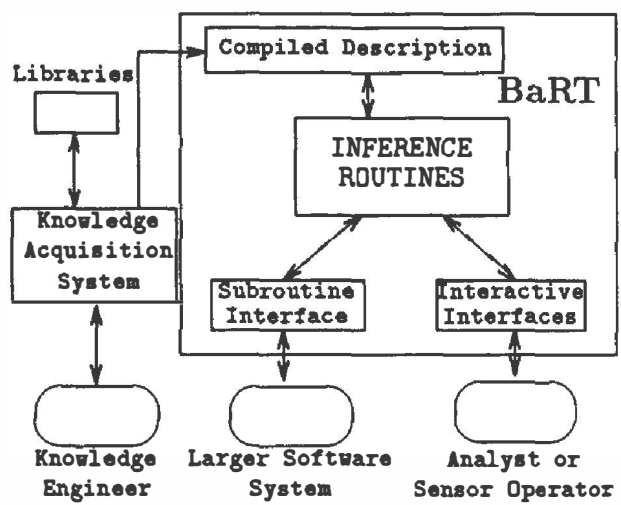

Figure 2: BaRT system architecture.

The BaRT compiler takes descriptions generated by the knowledge acquisition system and converts them into the form required by the core inference routines. The important point about the compiler is that it can identify conditions like loops that require special processing. The compiler can also identify canonical interactions like the OR gate for binary variables where fast methods (Pearl, 1988) can be substituted for general ones to make updating more efficient.

The inference methods in BaRT can be invoked using either a subroutine interface or an interactive window interface. The subroutine interface is used when BaRT performs inferences for some larger knowledge based system that selectively makes use of various belief maintenance capabilities (eg. Morawski (1989)). The interactive interface is used when BaRT itself is the primary problem solver (eg. Booker & Hota (1988)).

BaRT is written in Common Lisp using the CLOS object-oriented programming standard. For Lisp environments that do not support CLOS, BaRT provides a mini-CLOS package that uses defstructs to simulate the functionality BaRT requires. This means that BaRT can be run with the subroutine interface in any Common Lisp environment. Graphics interfaces have been developed for the Symbolics Lisp machine and the Sun workstation. An X-window interactive interface will be developed to make BaRT easier to port to other machines.

### 3.2. Knowledge Representation

The current version of BART supports three knowledge representations: Bayesian networks, influence diagrams, and taxonomic hierarchies. We are currently implementing a method for default reasoning and property inheritance that is based on a probabilistic semantics (Geffner, 1988). This will give BART the capability to represent many kinds of conceptual relationships and make categorical inferences in an axiomatic way. The fourth type of knowledge representation supported by BART will be databases containing this kind of information.

The basic inference procedure used in Bayesian networks is Pearl's (1986) distributed message passing algorithm for singly connected networks. We have implemented this algorithm using tensor products in a manner that is efficient and does not depend on any special properties of the link matrices (Booker, Hota, & Hemphill, 1989). An important advantage of Pearl's algorithm is that it is readily implemented on parallel hardware. A version of BART that exploits this property has been implemented on the Butterfly Plus Parallel Processor (Ramsey & Booker, 1988). Another advantage of this algorithm is that, with slight modification, it also computes a categorical belief commitment (Pearl, 1987) for every node in the network. This is useful for explanations and for making abductive inferences. BART converts networks with loops into singly connected networks using Chang and Fung's (1989) node aggregation algorithm.

BART solves influences diagram problems by using Cooper's method (1988) to convert the influence diagram into a belief network. Cooper's algorithm recursively constructs and evaluates all paths in the decision tree that corresponds to the influence diagram. The efficiency of the BART implementation is enhanced with a simple branch and bound technique. Before any path is explored, BART makes an optimistic assumption about the expected value that will be computed for that path. Whenever this optimistic bound is worse than the expected value of some fully expanded alternative, the candidate path is pruned.

Belief in taxonomic hierarchies is updated using another algorithm devised by Pearl (1986). BART implements the weight distribution/normalization version of this procedure in which the belief of each hypothesis is the sum of the beliefs of its constituent singleton hypotheses. This means that BART can propagate evidence in any class-subclass hierarchy.

Several auxiliary computations are also available in BART: auxiliary variables to compute Boolean constraints and queries, and an error-based measure of impact to gauge the potential effect an uninstantiated node might have on the uncertainty of a given target hypothesis (Pearl, 1988).

### 3.3. Relationship to CSRL

The relationship between BART's capabilities and the generic task view of classificatory problem solving is straightforward. A taxonomic hierarchy is the counterpart to the classification tree used in CSRL. The BART implementation has the advantage of allowing arbitrary subclass-superclass hierarchies to be defined. This avoids the knowledge engineering task of having to simplify the set of classificatory hypotheses into a tree, and allows more than one perspective on the hierarchical classification task to be considered at the same time.

Several methods are available to implement the function of knowledge groups. The data abstractions and categorical inferences required in the knowledge-directed information passing task can be managed with probabilistic techniques for default reasoning (Geffner, 1988). Hypothesis matching can be accomplished using taxonomies, influence diagrams, and Bayesian networks.

It is important to note that the qualitative rule-based computations used in CSRL's knowledge groups are really just a shorthand for describing the joint interaction between an evidential hypothesis and its specializations. In fact, the original formulation of knowledge groups used tables with qualitative entries to specify these interactions (Chandrasekaran & Tanner, 1986). BART will be augmented with a simple rule-based language for describing the local probabilistic interactions of nodes in a network. Each rule is an expression that can refer to the values of hypothesis variables and to qualitative terms having some predefined probabilistic interpretation. Once invoked, a rule assigns values to some subset of elements in the joint conditional probability matrix quantifying a network link. This is directly analogous to the CSRL knowledge group procedure, but it is accomplished without sacrificing semantic rigor.

Since all inferences computed by BART are probabilistic, the communication between a classificatory hypothesis and its knowledge group is also straightforward. The hypothesis simply instructs its associated knowledge group to acquire evidence and update itself. At present, the only control mechanism available for this evidence gathering activity is the impact measure for belief networks cited previously. Once the knowledge group acquires the data it needs and computes the resulting posterior beliefs, it returns a likelihood vector to the classificatory hypothesis. The likelihood vector then initiates an update of beliefs in the classification taxonomy. The variety of representations available in BaRT, coupled with a knowledge-based view of classification and the capability to explicitly model dependencies between hypothesis variables, distinguish BaRT from other Bayesian approaches to classification (eg. AutoClass (Cheeseman et al., 1988)).

An important area for further research is the development of control strategies for evaluating hypotheses in the taxonomy. It is easy to implement a strategy like simple establish-refine. What is desired is a method that takes cost-benefit considerations into account and allows for a wide variety of strategies to be specified. Influence diagrams have been proposed as a way to exercise decision-theoretic control of problem solving in other systems (Breese & Fehling, 1988). A similar approach might be suitable in BaRT.

## 4. Conclusion

The primary goal of the BaRT project is to make state of the art techniques for uncertain reasoning available to researchers concerned with classificatory problem solving. The thrust of our research has therefore been to design and implement a generic tool for hierarchical Bayesian reasoning. BaRT brings together several theoretical ideas about plausible inference in a way that appears to be efficient and practical for real applications. Preliminary versions of BaRT have been used as a decision aid for classifying ship images (Booker & Hota, 1988), and as the reasoning component of a system concerned with analyzing intelligence reports (Morawski, 1989).

When completed, BaRT will provide extensive facilities for building classificatory problem solvers. Specialized representations are available to handle each of the reasoning tasks associated with classificatory problem solving: taxonomic hierarchies for hierarchical classification, Bayesian networks and influence diagrams for hypothesis matching, and probabilistic default rules for knowledge-directed information passing. BaRT is comparable to a system like CSRL in that it allows a knowledge engineer to think in terms of the inherent, qualitative structure of a problem. Because all of BaRT's capabilities are based on sound probabilistic semantics, however, BaRT has the added advantage of computing normative and axiomatic inferences.

## Acknowledgements

The mini-CLOS package was written by Dan Hoey.